\documentclass[conference]{IEEEtran}
\IEEEoverridecommandlockouts
\usepackage{cite}
\usepackage{amsmath,amssymb,amsfonts}
\usepackage{algorithmic}
\usepackage{graphicx}
\usepackage{textcomp}
\usepackage{xcolor}
\def\BibTeX{{\rm B\kern-.05em{\sc i\kern-.025em b}\kern-.08em
    T\kern-.1667em\lower.7ex\hbox{E}\kern-.125emX}}

\usepackage[breaklinks=true,colorlinks,bookmarks=false]{hyperref}
\usepackage[nameinlink]{cleveref}
\usepackage[ruled,vlined]{algorithm2e} 

\usepackage{comment}
\usepackage{todonotes}

\crefname{algocf}{algo.}{algorithms}
\Crefname{algocf}{Algorithm}{Algorithms}
\definecolor{mediumseagreen}{rgb}{0.24, 0.7, 0.44}
\definecolor{darkgoldenrod}{rgb}{0.72, 0.53, 0.04}
\definecolor{mediumred-violet}{rgb}{0.73, 0.2, 0.52}

\begin{document}

\title{Improving Video Instance Segmentation by Light-weight Temporal Uncertainty Estimates
\thanks{This work is supported by Volkswagen Group Automation.}
}

\makeatletter
\newcommand{\linebreakand}{%
  \end{@IEEEauthorhalign}
  \hfill\mbox{}\par
  \mbox{}\hfill\begin{@IEEEauthorhalign}
}
\makeatother

\author{\IEEEauthorblockN{Kira Maag}
\IEEEauthorblockA{\textit{University of Wuppertal \& ICMD} \\
kmaag@uni-wuppertal.de} \\
\IEEEauthorblockN{Fabian H\"uger}
\IEEEauthorblockA{\textit{Volkswagen AG} \\
fabian.hueger@volkswagen.de}
\and
\IEEEauthorblockN{Matthias Rottmann}
\IEEEauthorblockA{\textit{University of Wuppertal \& ICMD} \\
rottmann@uni-wuppertal.de} \\
\IEEEauthorblockN{Peter Schlicht}
\IEEEauthorblockA{\textit{Volkswagen AG} \\
peter.schlicht@volkswagen.de}
\and
\IEEEauthorblockN{Serin Varghese}
\IEEEauthorblockA{\textit{Volkswagen AG} \\
john.serin.varghese@volkswagen.de} \\
\IEEEauthorblockN{Hanno Gottschalk}
\IEEEauthorblockA{\textit{University of Wuppertal \& ICMD} \\
hgottsch@uni-wuppertal.de}
}

\maketitle

\begin{abstract}
Instance segmentation with neural networks is an essential task in environment perception. In many works, it has been observed that neural networks can predict false positive instances with high confidence values and true positives with low ones. Thus, it is important to accurately model the uncertainties of neural networks in order to prevent safety issues and foster interpretability. In applications such as automated driving, the reliability of neural networks is of highest interest. In this paper, we present a time-dynamic approach to model uncertainties of instance segmentation networks and apply this to the detection of false positives as well as the estimation of prediction quality. The availability of image sequences in online applications allows for tracking instances over multiple frames. Based on an instances history of shape and uncertainty information, we construct temporal instance-wise aggregated metrics. The latter are used as input to post-processing models that estimate the prediction quality in terms of instance-wise intersection over union. The proposed method only requires a readily trained neural network (that may operate on single frames) and video sequence input. In our experiments, we further demonstrate the use of the proposed method by replacing the traditional score value from object detection and thereby improving the overall performance of the instance segmentation network. 
\end{abstract}

\section{Introduction}
Object detection describes the task of identifying and localizing objects of a set of given classes. With respect to image data, state-of-the-art approaches are mostly based on convolutional neural networks (CNNs). Localization can be performed, for example, by predicting bounding boxes or labeling each pixel that corresponds to a given instance. The latter is also known as instance segmentation (see \Cref{fig:iou_meta}) which is an essential tool for scene understanding and considered throughout this work. The reliability of neural networks in terms of prediction quality estimation \cite{DeVries2018,Maag2019,Rottmann2018} and uncertainty quantification \cite{Gal2016} is of highest interest, in particular in safety critical applications like medical diagnosis \cite{Ozdemir2017} and automated driving \cite{Le2018}. However, instance segmentation networks such as YOLACT \cite{Bolya2019} and Mask R-CNN \cite{He2017} do not give well adjusted uncertainty estimations \cite{Guo2017}. These networks provide a confidence value, also called score value, for each instance which can have high values for false predictions and low ones for correct predictions. Confidence calibration \cite{Guo2017} addresses this problem by adjusting the confidence values to reduce the error between the average precision (as a performance measure) and the confidence values \cite{Kueppers2020}. During inference of instance segmentation networks, all instances with score values below a threshold are removed. It can happen that correctly predicted instances disappear as well as many false positives remain. For this reason, we do not use a score threshold during inference and instead present an uncertainty quantification method that gives more accurate information compared to the simple score value. We utilize this uncertainty quantification to improve the networks' performance in terms of accuracy. Another approach to improve the trade-off between false negatives and false positives has been introduced in \cite{Chan2020}. Different decision rules are applied by introducing class priors which assign larger weight to underrepresented classes. 
\begin{figure}[t]
\center
    \includegraphics[width=0.485\textwidth]{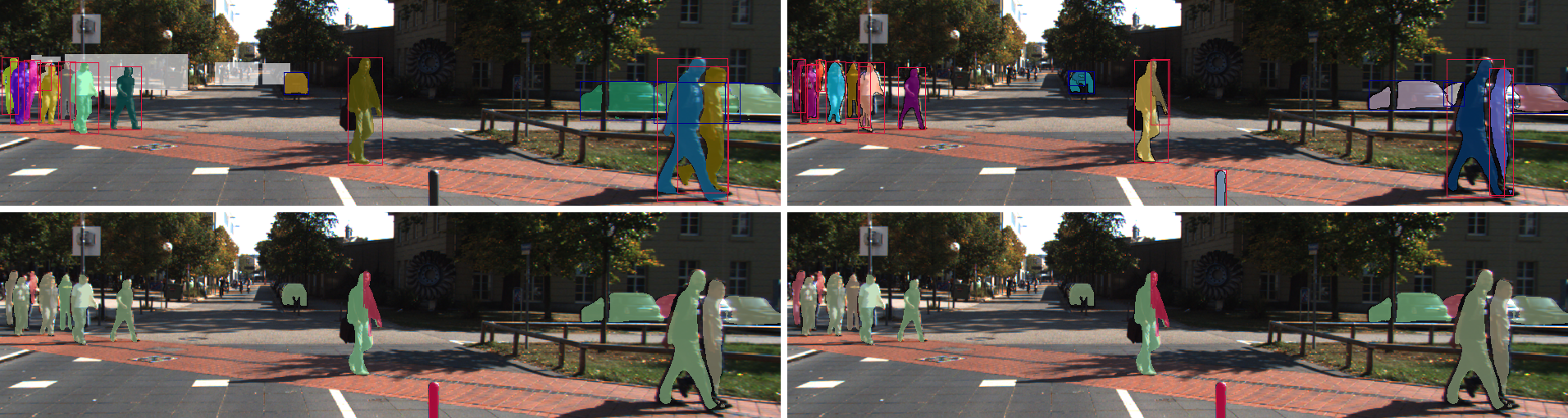}
    \caption{\emph{Top left}: Ground truth image with ignored regions (white). \emph{Top right}: Instance segmentation. In the first row, the bounding boxes drawn around the instances represent the class, blue denotes cars and red pedestrians. \emph{Bottom left}: A visualization of the true instance-wise IoU of prediction and ground truth. Green color corresponds to high IoU values and red color to low ones. \emph{Bottom right}: Instance-wise IoU prediction obtained from meta regression. }
    \label{fig:iou_meta}
\end{figure}

In this work, for the first time, we introduce the tasks of \emph{meta classification} and \emph{meta regression} for instance segmentation which was previously introduced only for semantic segmentation \cite{Rottmann2018}. The authors post-processed semantic segmentation predictions in order to estimate the quality of each predicted segment. Meta classification refers to the task of predicting whether a predicted instance intersects with the ground truth or not. A commonly used performance measure is the intersection over union (IoU) which quantifies the degree of overlap of prediction and ground truth \cite{Jaccard1912}. In instance segmentation, an object is called false positive if the IoU is less than $0.5$. Hence, we consider the task of (meta) classifying between IoU $< 0.5$ and IoU $\geq 0.5$ for every predicted instance. We use meta classification to identify false positive instances and improve the overall network performance compared to the application of a score threshold during inference. The task of meta regression is the prediction of the IoU for each predicted instance directly. Both meta classification and regression (\emph{meta tasks}) are able to reliably evaluate the quality of an instance segmentation obtained from a neural network. In addition, the prediction of the IoU serves as a performance estimate. For learning both meta tasks, we use instance-wise metrics as input for the respective models. In \cite{Maag2019} single frame metrics are introduced which characterize uncertainty and geometry of a given semantic segment. By tracking these segments over time, time series of single frame metrics are generated. We not only apply these metrics to instance segmentation, but also extend them by novel and truly time-dynamic metrics. These metrics are based on survival time analysis, on changes in the shape and on expected position of instances in an image sequence. From this information, we estimate the prediction quality on instance-level by means of temporal uncertainties. Additionally, for generating time series we propose a light-weight tracking approach for predicted instances. Our tracking algorithm matches instances based on their overlap in consecutive frames by shifting instances according to their expected location in the subsequent frame predicted via linear regression.

In this work, we present post-processing methods for uncertainty quantification and performance improvement. We only assume that a trained instance segmentation network and image sequences of input data are available. 
The source code of our method is publicly available at \url{https://github.com/kmaag/Temporal-Uncertainty-Estimates}. Our contributions are summarized as follows:
\begin{itemize}
    \item We present a light-weight tracking algorithm for instances predicted by a neural network and resulting time-dynamic metrics. These metrics then serve as input for different models for meta classification and meta regression.
    \item We evaluate our tracking algorithm on the KITTI \cite{Geiger2012} and the MOT dataset \cite{Milan2016}.
    \item We perform meta classification and regression to evaluate the quality of two state-of-the-art instance segmentation networks, the YOLACT and the Mask R-CNN network. Furthermore, we study different types of models for meta tasks w.r.t.\ their dependence on the length of time series and compare them with different baselines. For meta regression we obtain $R^2$ values of up to $86.85\%$ and for meta classification AUROC values of up to $98.78\%$ which is clearly superior to the performance of previous approaches.
    \item For the first time, we demonstrate successfully that time-dynamic meta classification performance can be traded for instance segmentation performance. We compare the meta classification performance with the application of a score threshold during inference. Meta classification reduces the number of false positives by up to $44.03\%$ while maintaining the number of false negatives.
\end{itemize}
The paper is structured as follows. In \Cref{sec:related_work} the related work on uncertainty estimation and object tracking methods is discussed. This is followed by the presentation of our method including the tracking algorithm, temporal metrics as well as meta classification and regression models in \Cref{sec:method}. Finally, we present the numerical results in \Cref{sec:result}.
%
%
\section{Related Work}\label{sec:related_work}
\subsection{False Positive Detection and Uncertainty Quantification} 
Bayesian models are one possibility to consider model uncertainty \cite{Mackay1992}. A well-known approximation to Bayesian inference is the Monte-Carlo (MC) Dropout \cite{Gal2016} which has proven to be practically efficient in detecting uncertainties and has also been applied to semantic segmentation tasks \cite{Lee2020}. In \cite{Wickstrom2018} MC Dropout is also used to filter out predictions with low reliability. To detect spatial and temporal uncertainty, this line of research is further developed in \cite{Huang2018}. Based on MC Dropout, structure-wise metrics are presented in \cite{Roy2018}, as well as voxel-wise uncertainty metrics based on softmax maximum probability in \cite{Hoebel2020}. In \cite{Rottmann2018} the concepts of meta classification and meta regression are introduced with segment-wise metrics as input extracted from the segmentation network's softmax output. This idea is extended in \cite{Schubert2020} to object detection, in \cite{Schubert2019} by adding resolution dependent uncertainty and in \cite{Maag2019} by a temporal component.
Similar works on a single object per image basis are introduced in \cite{DeVries2018} and \cite{Huang2016}, instead of hand crafted metrics they utilize additional CNNs. Some methods for uncertainty evaluation are transferred to object detection, such as MC Dropout \cite{Kraus2020}. False positive detection is presented in \cite{Ozdemir2017} based on MC Dropout and an ensemble approach. In \cite{Le2018} false positive objects obtain high uncertainties using two methods, loss attenuation and redundancy with multi-box detection. Based on a dropout sampling approach for object detection \cite{Miller2018}, the work presented in \cite{Morrison2019} investigates the semantic and spatial uncertainty in instance segmentation.

For instance segmentation only uncertainty estimation methods based on dropout sampling \cite{Morrison2019} exist so far. While MC Dropout \cite{Gal2016,Huang2018,Lee2020,Wickstrom2018} is based on multiple runs of a network, our method can be applied to any network as a post-processing step using only information of the network output. The work closest to ours \cite{Maag2019} provides time series of single frame metrics for semantic segments and in confidence calibration \cite{Kueppers2020}, only score values and bounding box positions are considered. We go beyond the measures presented in \cite{Maag2019} and introduce truly time-dynamic metrics that quantify uncertainties.
%
%
\subsection{Multi-Object Tracking} 
Tracking multiple objects in videos, in applications like automated driving, is an essential task in computer vision \cite{Milan2016}. The tracking task (association problem) in \cite{Zhu2018}, \cite{Son2017} and \cite{Zhao2018} is solved by dual matching attention networks, a CNN using quadruplet losses and a CNN based on correlation filters, respectively. The previously described algorithms use one model for the object detection and another for the association problem. In \cite{Wang2019} a shared model for both task is presented. A focus on the improvement of long-term appearance models is demonstrated in \cite{Kim2018} based on a recurrent network. Most works, including those described above, work with bounding boxes, while there are other methods using a binary segmentation mask representation of the object \cite{Wang2018}. In \cite{Aeschliman2010} and \cite{Payer2018}, segmentation and tracking are jointly solved using a pixel-level probability model and a recurrent fully convolutional network, respectively. To perform the detection, segmentation and tracking tasks simultaneously, the Mask R-CNN network is extended by a tracking branch \cite{Yang2019}, by 3D convolutions to incorporate temporal information \cite{Voigtlaender2019} and by a mask propagation branch \cite{Bertasius2019}. In contrast, the sub-problems of classification, detection, segmentation and tracking are treated independently in \cite{Luiten2019}. Another work for multi-object tracking is based on the optical flow and the Hungarian algorithm \cite{Bullinger2017}. The method introduced in \cite{Maag2019} works with semantic segments and is based on the overlap of segments in consecutive frames.

Machine learning is widely used in object tracking approaches as described above, while our tracking method is solely based on the degree of overlap of predicted instances. Following the tracking algorithm for semantic segments \cite{Maag2019}, we introduce a method to track instances. Hence, we are able to get away with an easy algorithm and less matching steps. The tracking performance was not evaluated in \cite{Maag2019}. However, we evaluate our tracking algorithm and compare it with the deep learning approach presented in \cite{Voigtlaender2019}. Note that, in contrast to tracking methods integrated into object detection or instance segmentation networks, our approach is independent of the network and serves as a post-processing step.
%
%
%
%
\section{Temporal Prediction Quality}\label{sec:method}
Instance segmentation is an extension of object detection. In both tasks, multiple bounding boxes with corresponding class affiliations are predicted. In instance segmentation, an additional pixel-wise mask representation is included. In both tasks, first a score threshold is used to remove objects with low scores, and thereafter a non-maximum suppression is applied to avoid multiple predictions for the same object. 
Our method is based on temporal information of these remaining instances. We track instances over multiple frames in video sequences and generate time-dynamic metrics. From this information we estimate the prediction quality (meta regression) on instance-level and predict false positive instances (meta classification), also in order to improve the networks' performance in terms of accuracy.
In this section, we describe our tracking approach, the temporal metrics as well as the methods used for meta classification and regression.
%
%
\subsection{Tracking Method for Instances}\label{sec:tracking_segments}
In this section, we introduce a light-weight tracking algorithm for predicted instances in image sequences where instance segmentation is available for each frame. As our method is a post-processing step, the tracking algorithm is independent of the choice of instance segmentation network. Each instance $i$ of an image $x$ has a label $y$ from a prescribed label space $\mathcal{C}$. In \Cref{fig:iou_meta} (top left) a ground truth image is shown. Therein, the white areas are ignored regions with unlabeled cars and pedestrians. An evaluation for predicted instances in these regions is not possible, therefore all instances where $80\%$ of their number of pixels are inside an ignored region are not considered for tracking and further experiments. Given an image $x$, $\hat{\mathcal I}_x$ denotes the set of predicted instances not covered by ignored regions. We match instances of the same class in consecutive frames as follows: Instances are matched according to their overlap or if their geometric centers are close. For this purpose, we shift instances according to their expected location in the next frame using information from previous frames. We use a linear regression on $t_{l}$ consecutive frames to match instances that are at least one and at most $t_{l}-2$ frames apart in temporal direction in order to account for flashing instances (temporary false negatives or occluded instances). We define the overlap of two instances $i$ and $j$ by 
\begin{equation}\label{eq:overlap}
    O_{i,j} = \frac{| i \cap j |}{| i \cup j |}
\end{equation}
and the geometric center of instance $i$ represented as pixel-wise mask in frame $t$ by
\begin{equation}\label{geom_center}
    \bar{i}_{t} = \frac{1}{|i|} \sum_{(z_{v}, z_{h}) \in i} (z_{v}, z_{h})
\end{equation}
where $(z_{v}, z_{h})$ describes the vertical and horizontal coordinate of pixel $z$. We denote by $\{ x_{t} : t=1,\ldots,T \}$ the image sequence with a length of $T$. Our tracking algorithm is applied sequentially to each frame $t = 1, \ldots, T$. The instances in frame $1$ are assigned with random IDs. Then, the ones in frame $t-1$ follow a tracking procedure to match with instances in frame $t$. To give priorities for matching, the instances are sorted by size and passed in descending order. A detailed description of how an instance $i \in \hat{\mathcal I}_{x_{t-1}}$ in frame $t-1$ is matched with an instance $j \in \hat{\mathcal I}_{x_{t}}$ in frame $t$ is described in \Cref{alg:track}.
\begin{algorithm}[!th]
  \caption{Tracking algorithm}
  \SetAlgoLined
  \DontPrintSemicolon
  \SetAlgoNoEnd
  /* shift */ \;
  \If{$t>2$ and instance $i$ exists in frame $t-2$}{
    shift instance $i$ from frame $t-1$ by the vector $ \left( \bar{i}_{t-1} - \bar{i}_{t-2} \right)$ \;
    \uIf{$\max_{j \in \hat{\mathcal I}_{x_{t}}} O_{i,j} \geq c_{o}$}{
        match instances $i$ and $j$
    }
    \uElseIf{$\min_{j \in \hat{\mathcal I}_{x_{t}}} \left\| \bar{j}_{t} - \bar{i}_{t-1} \right\|_{2} + \left\| \left( \bar{i}_{t-1} - \bar{i}_{t-2} \right) - \left( \bar{j}_{t} - \bar{i}_{t-1} \right)  \right\|_{2} \leq c_{d}$}{
            match instances $i$ and $j$
    }
  }
  /* distance */ \; 
  \If{$t>1$ and $i$ does not exist in frame $t-2$}{
    \If{$\min_{j \in \hat{\mathcal I}_{x_{t}}} \left\| \bar{j}_{t} - \bar{i}_{t-1} \right\|_{2} \leq c_{d}$}{
      match instances $i$ and $j$
    }
  }
  /* overlap */ \;
  \If{$t>1$ and $\max_{j \in \hat{\mathcal I}_{x_{t}}} O_{i,j} \geq c_{o}$}{
      match instances $i$ and $j$
  }
  /* regression */ \;
  \If{$t>3$ and instance $i$ appears in at least two of the frames $t-t_{l}, \ldots, t-1$}{
    compute geometric centers of instance $i$ in frames $t-t_{l}$ to $t-1$ (in case $i$ exists in all these frames) \;
    perform linear regression to predict the geometric center $(\hat{\bar{i}}_{t})$ \;
    \eIf{$ \min_{j \in \hat{\mathcal I}_{x_{t}}} \left\| \bar{j}_{t} - \hat{\bar{i}}_{t} \right\|_{2} \leq c_{l}$}{
      match instances $i$ and $j$
    }
    {  shift instance $i \in \hat{\mathcal I}_{x_{t_{max}}}$ by the vector $\left( \hat{\bar{i}}_{t} - \bar{i}_{t_{max}} \right)$ where $t_{max} \in \{t-t_{l}, \ldots, t-1\}$ denotes the frame where $i$ contains the maximum number of pixels \;
      \If{$\max_{j \in \hat{\mathcal I}_{x_{t}}} O_{i,j} \geq c_{o}$}{
        match instances $i$ and $j$
      }
    }
  }
\label{alg:track}
\end{algorithm}
This is performed for every instance $i$. If the instances $i$ and $j$ are matched, the algorithm terminates and instance $j$ is excluded from further considerations in order to preserve the instance uniqueness. When the algorithm has visited all instances $i \in \hat{\mathcal I}_{x_{t-1}}$, the instances $j \in \hat{\mathcal I}_{x_{t}}$ that have not been matched (e.g.\ newly occurring instances) are assigned with new IDs. Within the description of the tracking algorithm, we introduce parameters $c_{o}$, $c_{d}$ and $c_{l}$ as thresholds for overlap, distance and distance after shifting according to linear regression, respectively. 
%
%
\subsection{Temporal Instance-wise Metrics}\label{sec:metrics}
First, we consider the metrics introduced in \cite{Maag2019} applied to instances. These metrics are single frame metrics based on an object's geometry, like instance size and geometric center, as well as extracted from the segmentation network's softmax output, such as an average over an instance's pixel-wise entropy. To calculate instance-wise metrics from any uncertainty heatmap (like pixel-wise entropy), we compute the mean of the pixel-wise uncertainty values of a given instance. In addition, an instance is divided into inner and boundary. The ratio of pixels in the inner and the boundary indicates fractal shaped instances which signals a false prediction. We analogously define uncertainty metrics for the inner and boundary since uncertainties may be higher on an instance's boundary. To this end, for each pixel $z$ a probability distribution over the classes $y$ is required. If the network does not provide a probability distribution for each pixel, but only for the instance, we only use those metrics of \cite{Maag2019} that can be computed from the predicted instance mask. We denote both sets of metrics by $U^{i}$ independent of the network. These metrics serve as a baseline in our tests and as a basis for the following metrics.

Next, we define six additional metrics mostly based on temporal information extracted by the tracking algorithm. Instance segmentation as an extension of object detection provides for each instance a confidence value. We add this \emph{score value}, denoted by $s$, to our set of metrics.

The next metric is based on the variation of instances in consecutive frames. Instance $i$ of frame $t-1$ is shifted such that $i$ and its matched counterpart $j$ in frame $t$ have a common geometric center. We then calculate the overlap \eqref{eq:overlap} as a measure of \emph{shape preservation} $f$. Large deformations may indicate poorly predicted instances. 

In the following, time series of the previously shown metrics are constructed. For each instance $i$ in frame $t$ we gather a time series of the geometric centers $\bar{i}_{k}$, $k=t-5, \ldots, t-1$. If the instance exists in at least two previous frames, the geometric center $\hat{\bar{i}}_{t}$ in frame $t$ is predicted using linear regression. The \emph{deviation} between geometric center and expected geometric center $\| \hat{\bar{i}}_{t} - \bar{i}_{t} \|_{2}$ is used as a time-dynamic measure denoted by $d_{c}$. 
We proceed analogously with the instance size $S=|i|$ and compute the \emph{size deviation} $ |\hat{S} - S | =: d_{s}$. Small deviations $d_{c}$ and $d_{s}$ indicate that the predicted instance is consistent over time. 

The following metric is based on a survival analysis \cite{Moore2016} of the instances. All predicted instances that are matched in the previous five frames with the same ground truth instance are chosen. A predicted instance $i$ and a ground truth instance $g$ are considered as a match if they have an IoU $\geq 0.5$. The associated survival time of instance $i$ in frame $t$ is described by the number of frames in which ground truth instance $g$ appears in consecutive frames. By the proposed tracking method, we obtain time series for each of the previously presented single frame metrics $U^{i}_{k} \cup \{ s_{k} \}$, $k=t-5, \ldots, t$. These serve as input for a Cox regression \cite{Cox1972} survival model which predicts the \emph{survival times} $v$ of instances. A higher value of $v$ indicates reliable instances, while a lower value suggests uncertainty. 

The final measure is based on the height to width ratio of the instances. To this end, we calculate the average height to width ratio of ground truth instances $g_{c}$ per class $c$ by
\begin{equation}
    r^{\text{GT}}_{c} = \frac{1}{|g_{c}|} \sum_{g=1}^{g_{c}} \frac{h_{g}}{w_{g}}
\end{equation}
where $h_{g}$ denotes the height and $w_{g}$ the width of an instance $g$. We separate the ratio by class, as for instance cars and pedestrians typically have different ratios of height and width. False predictions can result in deviations of these typical ratios. The \emph{ratio metric} for a predicted instance $i$ is given by $r := (h_{i} / w_{i}) / r^{\text{GT}}_{c}$.

For the calculation of the ratio $r^{\text{GT}}_{c}$ and the training of the survival model, only the ground truth data of the training set is used. Thus, the metrics $r$ and $v$ for instances in the test dataset can be determined without knowledge of ground truth. In summary, we use the following set of metrics $V^{i} = \{ U^{i} \} \cup \{ s, f, d_{c}, d_{s}, v, r \}$. An overview of the metrics' construction is shown in \Cref{fig:bock}.
\begin{figure}
\center
    \includegraphics[width=0.46\textwidth]{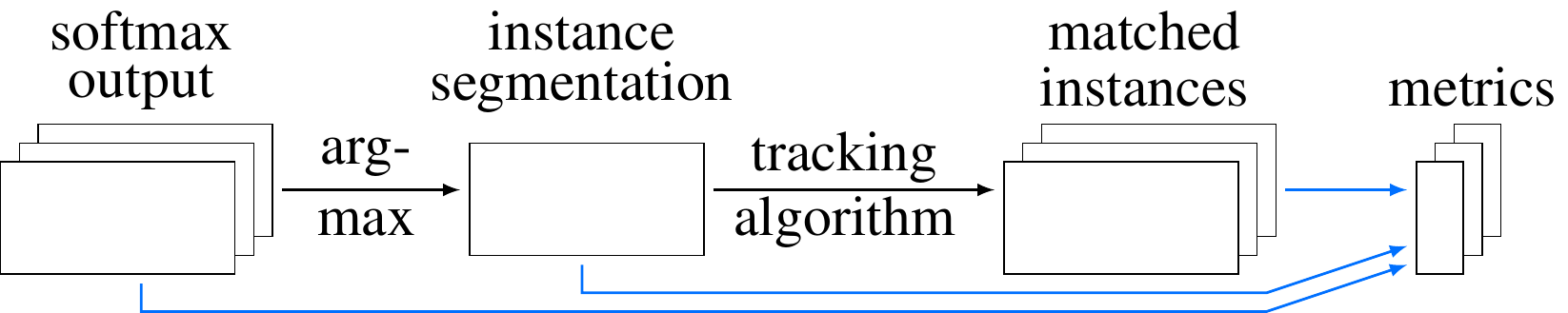}
    \caption{Overview of the metrics' construction that only requires a sequence of the softmax outputs. The information is extracted from the softmax output and further downstream from the instance segmentation and the instance tracking (\emph{blue arrows}). The resulting metrics serve as input for the meta tasks.}
    \label{fig:bock}
\end{figure}
%
%
\subsection{IoU Predictions}\label{sec:pred}
The intersection over union is a measure to determine the prediction accuracy of segmentation networks with respect to the ground truth. A predicted instance $i$ is matched with a ground truth instance $g$ if its overlap is the highest compared to the other ground truth instances. The IoU \eqref{eq:overlap} is then calculated between these two instances $i$ and $g$. In this work, we perform instance-wise predictions of the IoU (meta regression) comparing different regression approaches. In addition, we classify between IoU $< 0.5$ and IoU $\geq 0.5$ (meta classification) for all predicted instances. If the IoU of an instance is less than $50\%$, this instance is considered as a false positive. 
The metrics introduced in \Cref{sec:metrics} serve as input for both prediction tasks. As for the survival model, we compute time series of these metrics. We have for an instance $i \in \hat{\mathcal I}_{x_{t}}$ in frame $t$ the metrics $V^{i}_{t}$, as well as $V^{i}_{t'}$ from previous frames $t' < t$ due to object tracking. Meta classification and regression are performed by means of the metrics $V^{i}_{k}$, $k = t-n_{c}, \ldots, t$, where $n_{c}$ describes the number of considered frames. For regression and classification we use gradient boosting regression (\textcolor{mediumred-violet}{GB}) \cite{Friedman2002}, a shallow neural network containing only a single hidden layer with $50$ neurons (\textcolor{mediumseagreen}{NN L2}) and linear/logistic regression with $\ell_{1}$-penalization (\textcolor{darkgoldenrod}{LR L1}). Linear regression with $\ell_{1}$-penalization is also known as LASSO \cite{Tibshirani1996}. We analyze the benefit from using time series and the extent to which the additional metrics $V^{i} \setminus U^{i}$ yield improvements.
%
%
%
\section{Numerical Results}\label{sec:result}
In this section, we evaluate our light-weight tracking algorithm and investigate the properties of the metrics defined in the previous section. We perform meta regression and classification and study the influence of the length of the time series considered as well as different methods for the meta tasks. Furthermore, we study to which extent false positive detection can be traded for additional object detection performance and thus serve as an advanced score. To this end, we compare meta classification with ordinary score thresholds in terms of numbers of false positives and false negatives. We perform our tests on the KITTI dataset \cite{Geiger2012} for multi-object tracking and instance segmentation, which contains $21$ street scene videos from Karlsruhe (Germany) consisting of $1,\!242 \times 375$ $8,\!008$ images.
Additionally, we use the MOT dataset \cite{Milan2016} with scenes from pedestrian areas and shopping malls, which consists of $2,\!862$ images with resolutions of $1,\!920 \times 1,\!080$ ($3$ videos) and $640 \times 480$ ($1$ video).
For both datasets annotated videos are available \cite{Voigtlaender2019}. In contrast to the KITTI dataset, the MOT dataset only includes labels for the class pedestrian, but not for car.
In our experiments, we consider two different networks, the YOLACT network \cite{Bolya2019} and the Mask R-CNN \cite{He2017}.
The YOLACT network has a slim architecture designed for a single GPU. We retrain the network using a ResNet-50 \cite{He2015} backbone and starting from backbone weights for ImageNet \cite{Russakovsky2015}. We choose $12$ image sequences consisting of $5,\!027$ from the KITTI dataset (same splitting as in \cite{Voigtlaender2019}) and $300$ images from the MOT dataset for training. As validation set we use the remaining $9$ sequences of the KITTI dataset, achieving a mean average precision (mAP) of $57.06\%$. 
The Mask R-CNN focuses on high-quality instance-wise segmentation masks. As backbone we use a ResNet-101 and weights for COCO \cite{Lin2014}. We choose the same $12$ videos of the KITTI dataset as training set as well as $9$ videos as validation set achieving a mAP of $89.87\%$.
In the training of both networks, the validation set is neither used for parameter tuning nor early stopping. We choose a score threshold of 0 to use all predicted instances for further experiments. For instance tracking (see \Cref{alg:track}), we use the following values: $c_\mathit{o}=0.35$, $c_\mathit{d}=100$ and $c_\mathit{l}=50$. In our experiments, we use metrics extracted from the segmentation network’s softmax output. The Mask R-CNN outputs a probability distribution per pixel, while the YOLACT network only returns one probability per instance. For this reason, the set $U^{i}$ consists of more metrics for Mask R-CNN.
%
%
\subsection{Evaluation of our Tracking Algorithm}
For the evaluation of our tracking algorithm, we use common object tracking metrics, such as the multiple object tracking precision (MOTP) and accuracy (MOTA) \cite{Bernardin2018}. The MOTP is the averaged distance between geometric centers (geo) or bounding box centers (bb) of matched ground truth and predicted instances. The MOTA is based on three error ratios, the ratio of false negatives, false positives and mismatches ($\overline{\mathit{mme}}$). A mismatch error occurs when the ID of a predicted instance that was matched with a ground truth instance changes. In addition, we define by $\mathit{GT}$ all ground truth objects of an image sequence which are identified by different IDs and divide these into three cases, following \cite{Milan2016}. An object is mostly tracked ($\mathit{MT}$) if it is tracked for at least $80\%$ of frames (out of the total number of frames in which it appears), mostly lost ($\mathit{ML}$) if it is tracked for less than $20\%$, else partially tracked ($\mathit{PT}$). 

In \cite{Voigtlaender2019} the TrackR-CNN is presented which tracks objects and performs instance segmentation, both by means of a single neural network. This method is tested on the MOT dataset and the validation dataset of KITTI. To compare our approach with the TrackR-CNN method, we use the instances predicted by this network and apply our tracking algorithm to these instances to assess only the tracking quality (not the instance prediction quality of the network). The performance metrics mainly evaluate the object detection, while object tracking is only evaluated in the ratio of mismatches and MOTA. For this reason, we consider the latter performance metrics for our comparison on the TrackR-CNN instances. The results are given in \Cref{tab:track_ours}.
\begin{table}[t]
\centering
\caption{Mismatch ratio $\overline{\mathit{mme}}$ and MOTA results obtained by TrackR-CNN and by our tracking approach.}
\label{tab:track_ours}
\scalebox{0.95}{
\begin{tabular}{ccc|cc}
\cline{1-5}
\multicolumn{1}{c}{} & \multicolumn{2}{c|}{TrackR-CNN}  &  \multicolumn{2}{c}{ours} \\
\cline{1-5}
\multicolumn{1}{c}{} & $\overline{\mathit{mme}}$ & MOTA & $\overline{\mathit{mme}}$ & MOTA \\
\cline{1-5}
MOT & $0.0117$ & $0.6657$ & $0.0185$ & $0.6589$ \\
KITTI & $0.0153$ & $0.7993$ & $0.0235$ & $0.7911$ \\
\cline{1-5}
\end{tabular} }
\end{table}
We obtain for the MOT dataset a slightly higher mismatch ratio than TrackR-CNN and a MOTA value which is only $0.68$ percent points lower. For the KITTI dataset the results are similar. In summary, we are slightly weaker in tracking performance than TrackR-CNN, however, our algorithm does not use deep learning and is independent of the choice of instance segmentation network which enables us to conduct time-dynamic uncertainty quantification for any given instance segmentation network.

For further tests, we use YOLACT and Mask R-CNN for instance prediction. The results of object tracking metrics and performance measures for the KITTI dataset are shown in \Cref{tab:tracking_ym}.
\begin{table}[t]
\centering
\caption{Object tracking results and performance measures for the KITTI dataset.}
\label{tab:tracking_ym}
\scalebox{0.925}{
\begin{tabular}{cccccc}
\cline{1-6}
 & precision & recall & MOTP$_\text{bb}$ & MOTP$_\text{geo}$ & MOTA \\
\cline{1-6}
YOLACT & $0.3186$ & $0.7211$ & $3.34$ & $2.68$ & $-0.8746$ \\
Mask & $\mathbf{0.5546}$ & $\mathbf{0.9379}$ & $\mathbf{2.61}$ & $\mathbf{2.39}$ & $\mathbf{0.1281}$ \\
\cline{1-6}
 & $\mathit{GT}$ & $\mathit{MT}$ & $\mathit{PT}$ & $\mathit{ML}$ & $\overline{\mathit{mme}}$ \\
\cline{1-6}
YOLACT & $219$ & $124$ & $75$ & $20$ & $\mathbf{0.0539}$ \\
Mask & $219$ & $204$ & $13$ & $2$ & $0.0564$ \\
\cline{1-6}
\end{tabular} }
\end{table}
Mask R-CNN mainly achieves better results than YOLACT. This can be attributed to the fact that Mask R-CNN achieves higher mAP values than YOLACT. 
%
%
\subsection{Meta Classification and Regression}
While the KITTI dataset contains 10 frames per second (fps) videos of street scenes, the MOT dataset contains mostly 30 fps videos of pedestrian scenes. Due to slower motions, the data extracted from the images is very redundant and we get fewer different instances. This results in significant overfitting problems for the meta tasks. Also downsampling the frame rate is not an option due to the lack of data. Hence, we only consider the KITTI dataset for further experiments.

For meta classification (false positive detection: IoU $<0.5$ vs.\ IoU $\geq0.5$) and meta regression (prediction of the IoU), we use the KITTI validation set consisting of $2,\!981$ images. In our tests, we choose relatively low score thresholds, i.e., $0.1$ for YOLACT and 0.4 for Mask R-CNN. This is done to balance the number of false positives, such that the meta tasks obtain enough training data and the tracking performance is not significantly degraded. In order to investigate the predictive power of the metrics, we compute the Pearson correlation coefficients $\rho$ between the metrics $V^{i} \setminus U^{i}$ and the IoU (\Cref{tab:corr}).
\begin{table}[t]
\centering
\caption{Correlation coefficients $\rho$ with respect to IoU.}
\label{tab:corr}
\scalebox{0.915}{
\begin{tabular}{ccccccc}
\cline{1-7}
 & $s$ & $f$ & $d_{c}$ & $d_{s}$ & $v$ & $r$ \\ 
\cline{1-7}
YOLACT & $0.8760$ & $0.8033$ & $0.4106$ & $0.2755$ & $0.8141$ & $0.7415$ \\ 
Mask & $0.7303$ & $0.6033$ & $-0.1082$ & $0.1199$ & $0.2049$ & $-0.1007$ \\
\cline{1-7}
\end{tabular} }
\end{table}
The score value $s$ as well as the shape preservation metric $f$ show a strong correlation with the IoU for both networks. For YOLACT the survival metric $v$ and the ratio $r$ demonstrate high correlations.  
For meta tasks, we use a combination of metrics and further investigate their predictive power as well as the influence of time series. First, we only present the instance-wise metrics $V^{i}_{t}$ of a single frame $t$ to the meta classifier/regressor, secondly we extend the metrics to time series with a length of up to $10$ previous time steps $V^{i}_{k}$, $k=t-10, \ldots, t-1$. For the presented results, we apply a (meta) train/validation/test splitting of $70\%/10\%/20\%$ and average the results over $10$ runs obtained by randomly sampling the splitting. The corresponding standard deviations are provided. 

For the YOLACT network, we obtain roughly $13,\!074$ instances (not yet matched over time) of which $4,\!486$ have an IoU $< 0.5$. For the Mask R-CNN this ratio is $17,\!211 / 6,\!614$. For meta classification, we consider as performance measures the classification accuracy and AUROC. For meta regression, we compute the standard errors $\sigma$ and $R^2$ values.
The best results (over the course of time series lengths) for meta classification and meta regression are given in \Cref{tab:yolact_mask}.
\begin{table*}[t]
\centering
\caption{Results for meta classification and regression for the different meta classifiers and regressors. The super script denotes the number of frames where the best performance and in particular the given values are reached.}
\label{tab:yolact_mask}
\scalebox{0.93}{
\begin{tabular}{cccc|ccc}
\cline{1-7}
\multicolumn{1}{c}{} & \multicolumn{3}{c|}{YOLACT}  &  \multicolumn{3}{c}{Mask R-CNN} \\
\cline{1-7}
\multicolumn{1}{c}{} & $\textbf{\textcolor{darkgoldenrod}{LR L1}}$  &  $\textbf{\textcolor{mediumred-violet}{GB}}$ &$\textbf{\textcolor{mediumseagreen}{NN L2}}$ & $\textbf{\textcolor{darkgoldenrod}{LR L1}}$  &  $\textbf{\textcolor{mediumred-violet}{GB}}$ &$\textbf{\textcolor{mediumseagreen}{NN L2}}$\\
\cline{1-7}
\multicolumn{1}{c}{} & \multicolumn{6}{c}{Meta Classification IoU $<0.5,\geq 0.5$} \\
\cline{1-7}
ACC & $90.22\%\pm3.06\%^2$ & $\mathbf{92.62}\boldsymbol{\%}\pm2.48\%^9$ & $90.49\%\pm2.34\%^1$ & $93.43\%\pm1.78\%^4$ & $\mathbf{95.07}\boldsymbol{\%}\pm1.24\%^5$ & $94.31\%\pm1.65\%^4$\rule{0mm}{3.5mm} \\
\cline{1-7}
AUROC & $95.01\%\pm1.60\%^6$ & $\mathbf{96.98}\boldsymbol{\%}\pm0.87\%^9$ & $95.04\%\pm1.19\%^3$ & $98.26\%\pm0.86\%^3$ & $\mathbf{98.78}\boldsymbol{\%}\pm0.53\%^6$ & $98.42\%\pm0.68\%^6$ \rule{0mm}{3.5mm} \\
\cline{1-7}
\multicolumn{1}{c}{} & \multicolumn{6}{c}{Meta Regression IoU} \\ 
\cline{1-7}
$\sigma$ & $0.164\pm0.013^7$ & $\mathbf{0.130}\pm0.018^5$ & $0.141\pm0.021^2$ & $0.170\pm0.017^8$ & $\mathbf{0.142}\pm0.020^{11}$ & $0.148\pm0.025^4$ \rule{0mm}{3.5mm} \\
\cline{1-7}
$R^2$ & $67.34\%\pm4.10\%^7$ & $\mathbf{79.87}\boldsymbol{\%}\pm2.66\%^5$ & $75.89\%\pm5.19\%^2$ & $81.36\%\pm3.76\%^8$ & $\mathbf{86.85}\boldsymbol{\%}\pm3.96\%^{11}$ & $85.84\%\pm3.94\%^4$ \rule{0mm}{3.5mm} \\
\cline{1-7}
\end{tabular} }
\end{table*}
Gradient boosting shows the best performance in comparison to linear models and neural networks with respect to all classification and regression measures. For meta classification, we achieve AUROC values of up to $98.78\%$. For meta regression, the highest $R^2$ value of $86.85\%$ for the Mask R-CNN network is obtained by incorporating $10$ previous frames. For this specific case, an illustration of the resulting quality estimate is shown in \Cref{fig:iou_meta}. We also provide video sequences\footnote{\url{https://youtu.be/6SoGmsAarTI}} that visualize the IoU prediction and instance tracking.
In \Cref{fig:timeline_reg_clas} (left) results for regression $R^2$ as functions of the number of frames are given. 
\begin{figure*}[t]
    \center
    \includegraphics[width=0.84\textwidth]{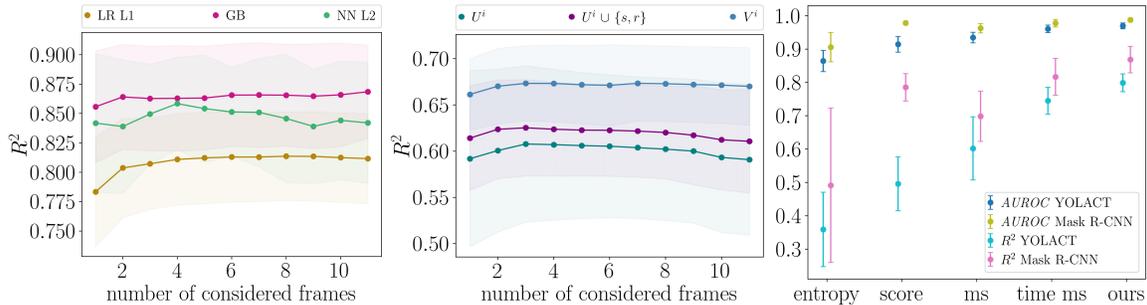}
    \caption{\emph{Left}: Results for meta regression $R^2$ as functions of the number of frames for the Mask R-CNN. \emph{Center}: Meta regression via linear regression with $\ell_{1}$-penalization for various input metrics and for the YOLACT network. \emph{Right}: Different baselines for both networks comparing AUROC and $R^2$ values. From left to right: mean entropy, score value, single frame MetaSeg (ms) approach \cite{Rottmann2018} with linear models, the time-dynamic extension \cite{Maag2019} using metrics $U^{i}$ and the best performing meta model, our method.}
    \label{fig:timeline_reg_clas}
\end{figure*}
We observe that the methods benefit from temporal information, although there are significant differences between them. 
In \Cref{fig:timeline_reg_clas} (right), we compare our best results for meta classification and regression for both networks with the following baselines. The results in \cite{Maag2019} were compared with a single-metric baseline as the mean entropy. We apply as single-metric also the mean entropy per instance as well as the score value using single frame gradient boosting. In addition, the approach in \cite{Rottmann2018} can be considered as a baseline. For this, we only apply the metrics $U^i$ for a single frame and the linear models. For the time-dynamic extension \cite{Maag2019} as baseline, the metrics $U^i$ are also used, however as time series and as input for Mask R-CNN and YOLACT. The best results for each method are displayed in \Cref{fig:timeline_reg_clas} (right). We observe that our method outperforms all baselines.
In \Cref{fig:timeline_reg_clas} (center) a comparison is shown between metrics $U^i$, all single frame metrics ($U^i$ plus score and ratio) and all metrics including the temporal ones. As shown before, metrics $U^i$ are clearly outperformed. The single frame metrics obtain significantly lower $R^2$ values compared to all metrics $V^i$. This difference can be observed when applying linear models as well as neural networks, whereas it is rather small when using gradient boosting. Gradient boosting has a strong tendency to overfitting and due to the small amount of data, we suspect that the gap between the performance of different metrics would also increase with more data.
In summary, we outperform all baselines by using our time-dynamic metrics as input for meta classifiers/regressors. We reach AUROC values of up to $98.78\%$ for classifying between true and false positives. 
%
%
\subsection{Advanced Score Values} \label{adv_score_values} 
In object detection, the score value describes the confidence of the network's prediction.
During inference, a score threshold removes false positives. If the threshold is raised to a higher value, not only false positives are removed, but also true positives. We study the network's detection performance while varying the score threshold. In total, we select $30$ different score thresholds from $0.01$ to $0.98$. Meta classification provides a probability of observing a false positive given a predicted instance. We threshold on this probability also with $30$ different thresholds and compare this to ordinary score thresholding. To this end, we feed gradient boosting as meta classifier with all metrics $V^{i}$ including $5$ previous frames. In \Cref{fig:fp_fn} the performance is stated in terms of the number of remaining false positives and false negatives. Each point represent one of the chosen thresholds.
\begin{figure}[t]
\center
    \includegraphics[width=0.49\textwidth]{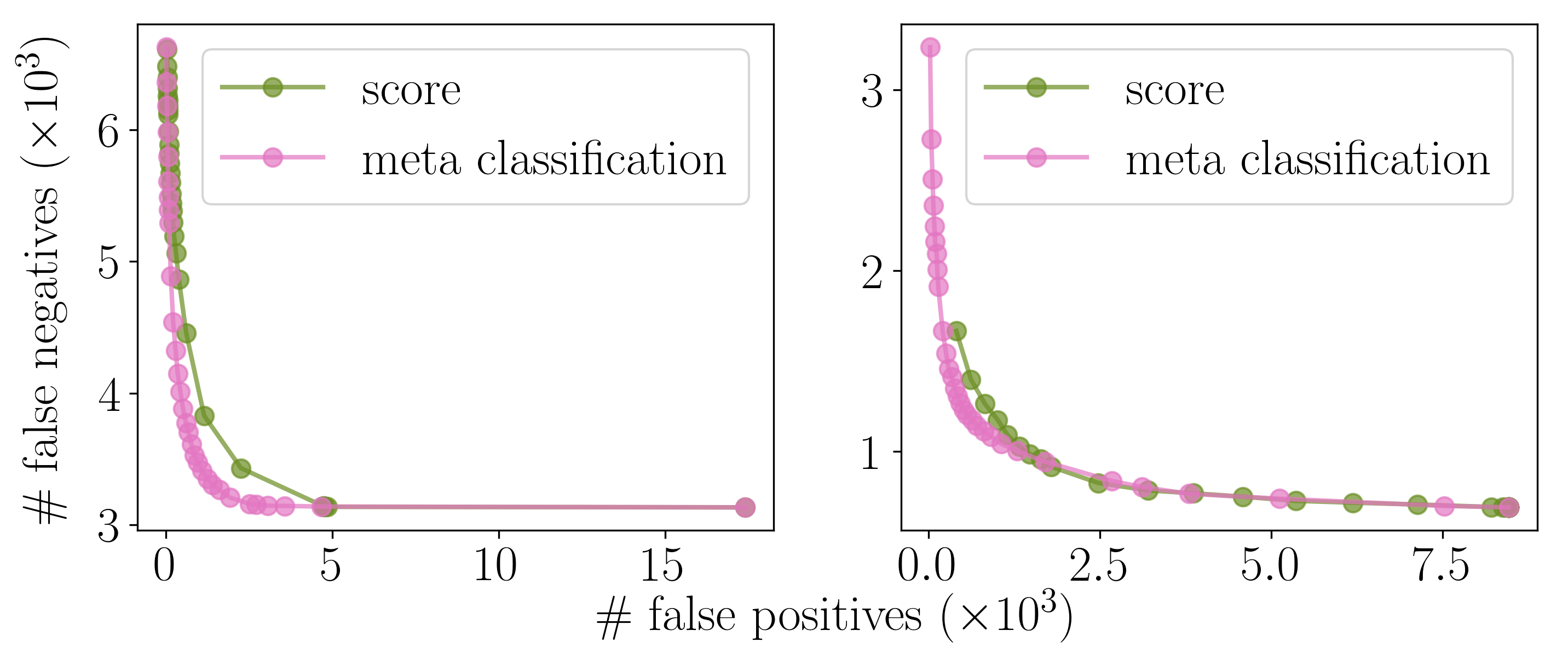}
    \caption{\emph{Left}: Number of false positive vs.\ false negative instances for different thresholds for the YOLACT network. \emph{Right}: Results for the Mask R-CNN.}
    \label{fig:fp_fn}
\end{figure}
For the YOLACT network, the meta classification achieves a lower number of errors, i.e., less false positives and false negatives. In comparison to score thresholding, we can reduce the number of false positives by up to $44.03\%$ for the approximate same number of false negatives. This performance increase is also reflected in the mAP. The highest mAP value obtained by score thresholding is $57.04\%$ while meta classification achieves $58.22\%$. 
When applying the Mask R-CNN, we can reduce the number of false positives up to $43.33\%$ for the approximate same number of false negatives. The maximum mAP values for both methods are similar. Score thresholding achieves an mAP of up to $89.87\%$ and meta classification of up to $89.83\%$. For both networks, we can improve the networks' performance by reducing false positives while the number of false negatives remains almost unchanged. Our method can be applied after training an instance segmentation network and therefore does not increase the network complexity. Up to some adjustment in the considered metrics, this approach is applicable to all instance segmentation networks.
%
%
\section{Conclusion}\label{sec:conclusion}
In this work, we proposed post-processing methods for instance segmentation networks, namely meta classification and meta regression. These methods are based on temporal information and can be used for both uncertainty quantification and accuracy improvement. We introduced a light-weight tracking algorithm for instances that is independent of the instance segmentation network. From tracked instances we constructed time-dynamic metrics (time series) and used these as inputs for the meta tasks. In our tests, we studied the influence of various time series lengths and different models on the meta tasks. For meta classification AUROC we obtain values of up to $98.78\%$ and for meta regression $R^2$ values of up to $86.85\%$ which is clearly superior to the performance of the baseline methods. Using meta classification we also improved the networks' prediction accuracy by replacing the score threshold by the estimated probability of correct classification during inference. We can reduce the number of false positives by up to $44.03\%$ while maintaining the number of false negatives.


\bibliographystyle{IEEEtran}
\bibliography{biblio}

\end{document}